\documentclass[conference]{IEEEtran}
\usepackage{cite}

\ifCLASSINFOpdf
   \usepackage[pdftex]{graphicx}
   \graphicspath{{figures/}}
\else
   \usepackage{graphicx}
\fi
\usepackage{amsmath}
\newcommand{\be}{\begin{equation}}
\newcommand{\ee}{\end{equation}}
\usepackage{array}
\usepackage{multirow}
\usepackage{booktabs}
\newcolumntype{L}[1]{>{\raggedright\let\newline\\\arraybackslash\hspace{0pt}}m{#1}}
\newcolumntype{C}[1]{>{\centering\let\newline\\\arraybackslash\hspace{0pt}}m{#1}}
\newcolumntype{R}[1]{>{\raggedleft\let\newline\\\arraybackslash\hspace{0pt}}m{#1}}

\ifCLASSOPTIONcompsoc
  \usepackage[caption=false,font=normalsize,labelfont=sf,textfont=sf]{subfig}
\else
  \usepackage[caption=false,font=footnotesize]{subfig}
\fi
\begin{document}
%
\title{Classification of Intra-Pulse Modulation of Radar Signals by Feature Fusion Based \\ Convolutional Neural Networks}

\author{\IEEEauthorblockN{Fatih Cagatay Akyon\IEEEauthorrefmark{1}\IEEEauthorrefmark{2},
		Yasar Kemal Alp \IEEEauthorrefmark{1},
		Gokhan Gok\IEEEauthorrefmark{1}\IEEEauthorrefmark{2},
		Orhan Arikan\IEEEauthorrefmark{2},
	}
	\IEEEauthorblockA{\IEEEauthorrefmark{1}Radar Electronic Warfare and Intelligence Systems Division, ASELSAN A.S., Ankara, Turkey}
	\IEEEauthorblockA{\IEEEauthorrefmark{2}Electrical and Electronics Engineering Department, Bilkent University, Ankara, Turkey}
	\{fcakyon,ykalp,gokhangok\}@aselsan.com.tr, \{oarikan\}@ee.bilkent.edu.tr}
\maketitle

\begin{abstract}
Detection and classification of radars based on pulses they transmit is an important application in electronic warfare systems. In this work, we propose a novel deep-learning based technique that automatically recognizes intra-pulse modulation types of radar signals. Re-assigned spectrogram of measured radar signal and detected outliers of its instantaneous phases filtered by a special function are used for training multiple convolutional neural networks. Automatically extracted features from the networks are fused to distinguish frequency and phase modulated signals. Simulation results show that the proposed FF-CNN (Feature Fusion based Convolutional Neural Network) technique outperforms the current state-of-the-art alternatives and is easily scalable among broad range of modulation types.
\end{abstract}


%
\IEEEpeerreviewmaketitle

\section{Introduction}
Automatic intra-pulse modulation recognition plays a pivoted role in radar classification systems \cite{lunden2007automatic}. Various methods are proposed to classify different intra-pulse modulations. Most of these methods are based on two major phases — feature extraction and classification. The classification phases do not vary much while methods are predominantly differentiated by their differences in the feature extraction phase.

Before the emergence of the convolutional neural network (CNN) solution, various signal processing methods are employed in feature extraction step for the differentiation between various intra-pulse modulation classes. In \cite{zeng2011automatic,zeng2012automatic,mingqiu2010classification}, and \cite{konopko2015radar} features are derived based on time-frequency analysis, and in \cite{rigling2010acf} and \cite{wang2016radar} features are extracted through autocorrelation functions. Apart from these methods, principal component analysis is performed in \cite{yu2009radar} and entropy method is applied in \cite{li2014radar}. For the classification phase, common machine learning methods are directly employed to classify extracted features. In \cite{lunden2007automatic}, artificial neural networks are employed. Support vector machines are used in \cite{ren2008radar} and \cite{mingqiu2009radar}. Clustering techniques are used in \cite{mingqiu2010classification} and \cite{yu2009radar}, and probabilistic graphical models are adopted in \cite{wang2016radar}.

The major weakness of aforementioned standard 2-phase techniques of feature extraction and classification, is that it is hard to extract features which facilitate classification. To overcome these weaknesses, a simple CNN based approach has been employed in  \cite{wang2017automatic}. In this method, feature extraction and classification are performed on a single network, yielding the highest performance and scalability reported to date. However, this method is mostly evaluated for frequency modulated signals and its classification performance on the some of the phase modulated pulses has not been investigated. Moreover, all previously done research focuses on low SNR levels up to -10 dB with the assumption that pulses are detected prior to the classification, which is not realistic considering it is far below the typical lowest SNR value for real-time pulse detection of EW receivers.

In this work, to overcome shortcomings of the previously mentioned techniques, we propose a feature fusion based convolutional neural network model (FF-CNN), that is capable of automatically performing feature extraction and classification of any type of frequency or phase modulated pulses. In the proposed technique, a previously detected radar pulse is first pre-processed to obtain a frequency and a phase related input. Then, the resultant data is input to a combined deep network structure composed of two CNNs followed by feature fusion layer that fuses the outputs of two independent CNNs. Such a feature fusion has been applied with significant success on other problems \cite{hang2016matrix, haghighat2016discriminant, pong2014multi, bai2015object}. Finally, the class probabilities are observed at the output. 

The details of pre-processing and proposed CNN model are covered in Chapter II. Simulation results are presented in Chapter III. The conclusions are drawn in chapter IV.

\section{Proposed FF-CNN Technique}
\begin{figure*}[t]
	\vspace{2ex}
	\begin{center}
		{\includegraphics[width=6.7in]{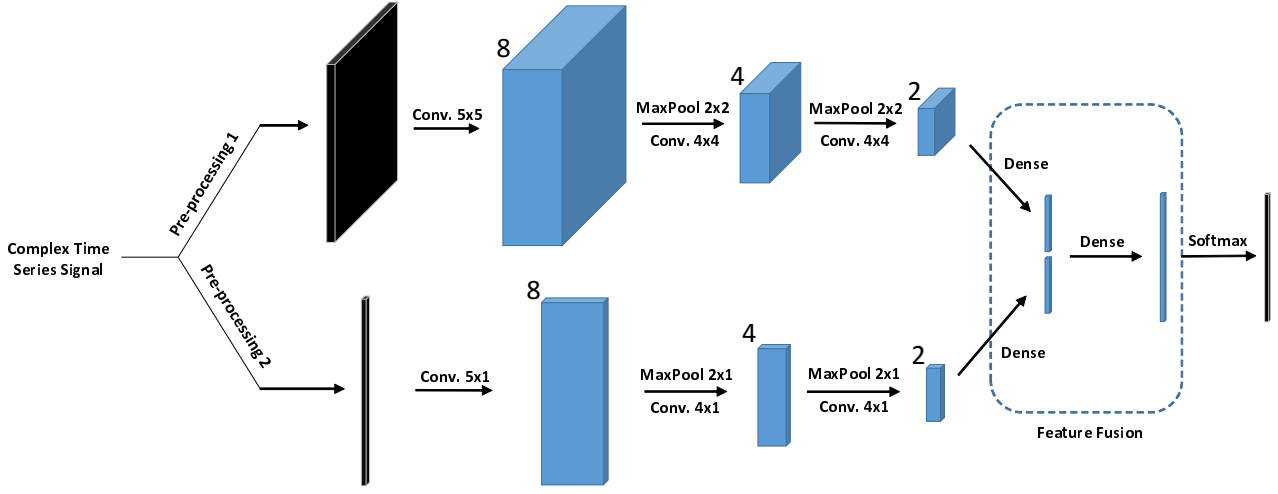}\label{fig:a1}}
		
		\caption{The proposed feature fusion based convolutional neural network (FF-CNN) model. First, two pre-processed inputs are subjected to feature extraction procedure through convolutional neural networks, then two network outputs are simultaneously combined and applied to fusion layers, and finally, softmax layer provides class probabilities.}
		\label{fig:model}
	\end{center}
\end{figure*}
Detected noisy radar pulse $x(t)$ can be modelled as follows:
\be
x(t) = a(t)e^{j\phi(t)} + z(t)
\ee
where $a(t)$ denotes the pulse envelope, $\phi(t)$ denotes instantaneous signal phase and $z(t)$ denotes zero mean circularly symmetric complex Gaussian noise. Two different pre-processing procedures are applied before network in order to facilitate both frequency and phase modulation identification of $x(t)$. This approach is different than traditional learning based methods with handcrafted features, since in FF-CNN these 2 automatically generated inputs are used in a network in an end-to-end manner. In other words, feature learning and classification is performed automatically.
First processing extracts Time-Frequency Images (TFIs) of the time-series complex signals which are good for differentiating frequency modulations. However, psuedo-random sequenced phase modulations have very similar TFIs. Thus, the second preprocessing is employed that makes the discrimination of phase modulated signals easier. 
Below, the pre-processing technique and proposed deep network structure are detailed.
\subsection{Pre-processing Stages}
\begin{figure}[t]
	\centering
	{\subfloat[]{\includegraphics[width=3.6in]{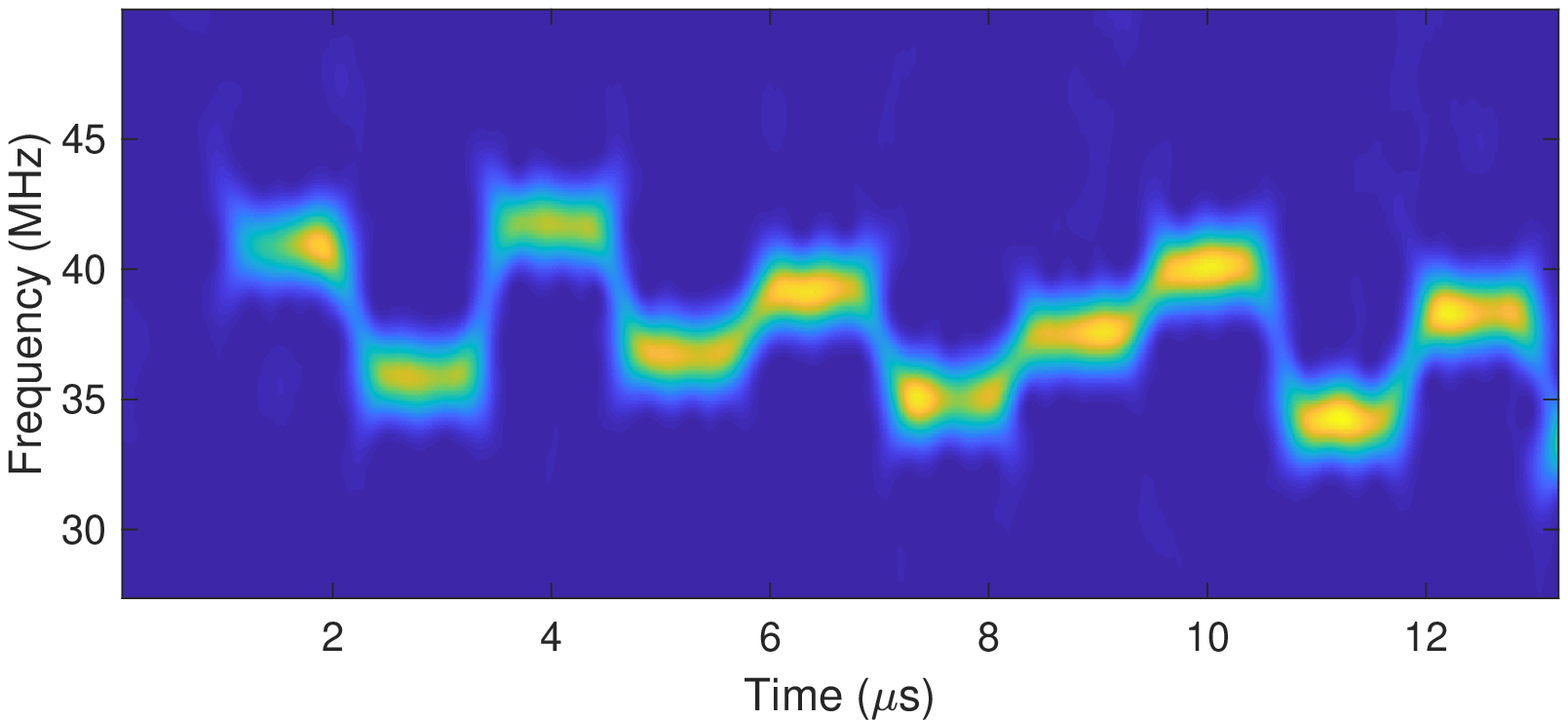}\label{fig:stft}}
		\vspace{-0.15in}
		
		\subfloat[]{\includegraphics[width=3.6in]{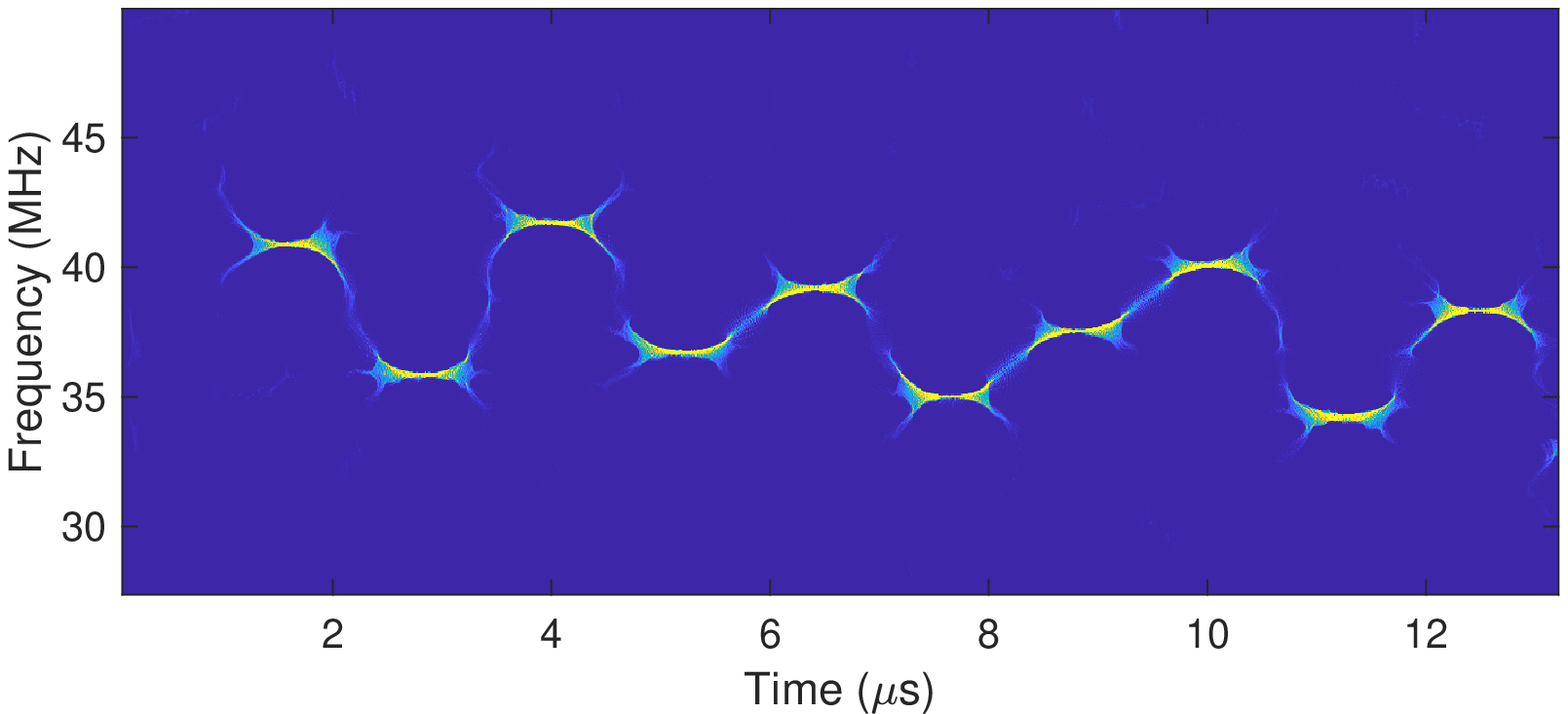}\label{fig:reas}}
		\vspace{0.05in}
		
		
		\caption{TFI’s of a Costas-10 modulated pulse at 10dB SNR using (a) STFT, and (b) RSTFT at 100 MHz sampling frequency.} 
		\label{fig:reas_stft}}
\end{figure}

In the first stage of pre-processing Reassigned Short-Time Fourier Transform (RSTFT) \cite{auger1995improving} of $x(t)$ is computed to generate high-resolution TFI of $x(t)$ to emphasize frequency modulations. Let $F_x(t,w;z)$ denote the STFT of $x(t)$, given as:
\begin{align}
F_x(t,w;z) = \int_{-\infty}^{\infty} x(s)z^*(s-t)e^{-jws} ds
\label{eq:ksfd}
\end{align}
where $z(t)$ is the windowing function controlling the desired time and frequency resolution of the resulting TFI. Then, RSTFT of the detected signal $x(t)$ is computed as:
\begin{align}
S_x^{r}(t',w') &= \int_{-\infty}^{\infty}\int_{-\infty}^{\infty} S_x(t,w) \delta(t'-\hat{t}_x(t,w))\\ &\hspace{0.7cm}\times\delta(w'-\hat{w}_x(t,w)) dtdw \nonumber
\label{eq:yenidenatama}
\end{align}
where $\hat{t}_x(t,w)$, $\hat{w}_x(t,w) $, and $S_x(t,w)$ are defined as:
\begin{align}
S_x(t,w)&= |F_x(t,w;z)|^2\\ 
\hat{t}_x(t,w)& = t - Re\left\{\frac{F_x(t,w;T_z(t))F_x^*(t,w;z)}{S_x(t,w)}\right\} \\
\hat{w}_x(t,w)& = w + Im\left\{\frac{F_x(t,w;D_z(t))F_x^*(t,w;z)}{S_x(t,w)}\right\}
\label{eq:ksfd1}
\end{align}
with $T_z(t)=tz(t)$ and $D_z(t)=\frac{dz(t)}{dt}$. Fig. \ref{fig:reas_stft} illustrates the STFT (\ref{fig:stft}) and the RSTFT (\ref{fig:reas}) of a frequency modulated $x(t)$ measured at 10 dB SNR. As demonstrated, the RSTFT provides a higher resolution TFI than STFT. However, since the high resolution TFI's are spatially sparse, they are downsampled to $128\times256$ by the nearest-neighbor interpolation method with a neglegible information loss \cite{parker1983comparison} to train the FF-CNN on a standardized input size with decreased training duration.
\begin{figure}[t]
	\centering
	{\subfloat[]{\includegraphics[width=3.6in]{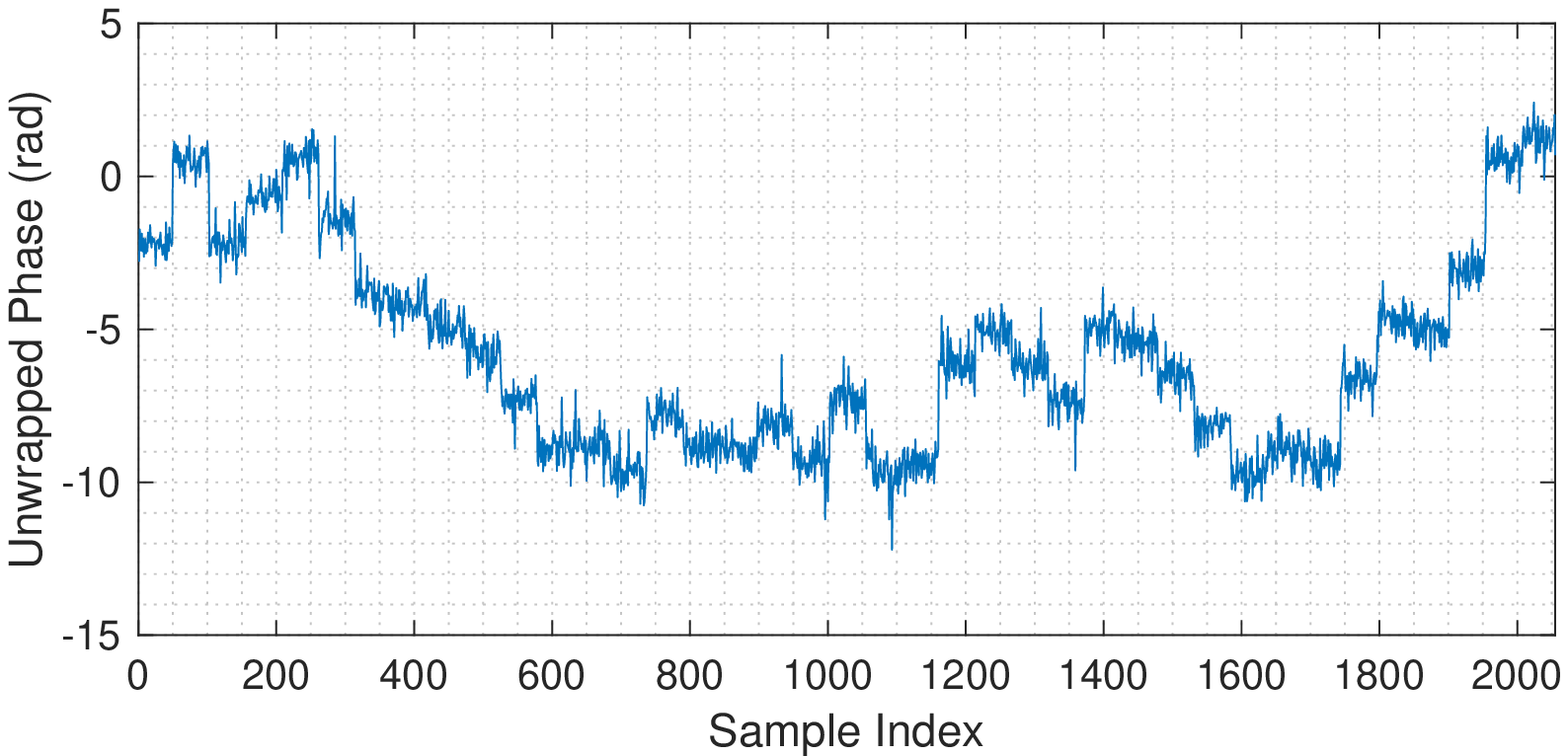}\label{fig:pp2_1}}
		\vspace{-0.15in}
		
		\subfloat[]{\includegraphics[width=3.6in]{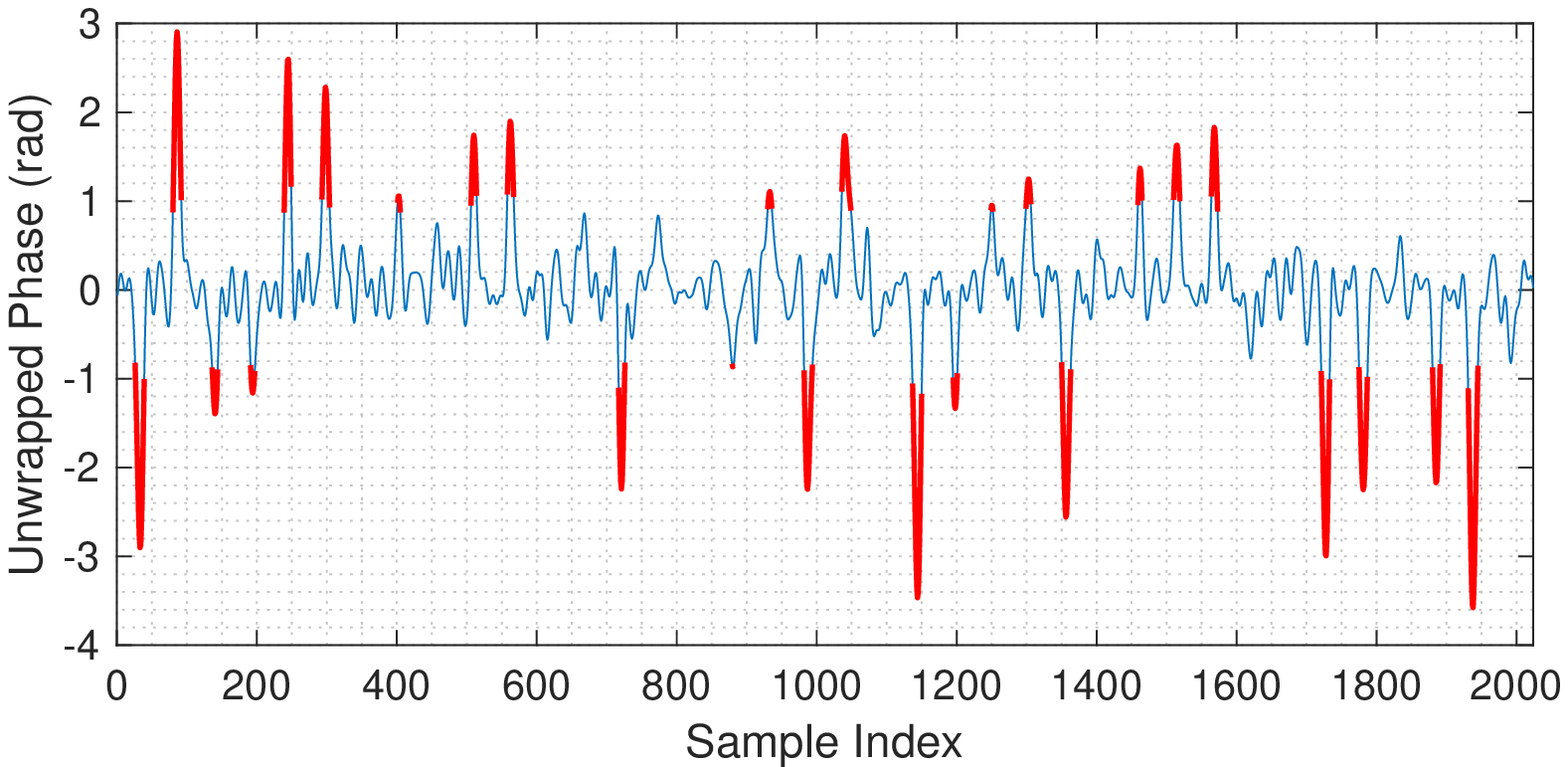}\label{fig:pp2_2}}
		\vspace{0.05in}
		
		\caption{Second pre-processing steps for a 16-PSK (phase) modulated pulse at 5 dB SNR. (a) Phase of the modulated signal, detected by applying a threshold. (b) Convolution of the pulse phase with HG (blue), detected phase jumps by robust least squares (red).} 
		\label{fig:pp2}}
\end{figure}

In the second stage of pre-preprocessing, first the unwrapped instantaneous phase of the measured signal is convolved with n=1 order HG (Hermit Gaussian) as:
\begin{align}\label{denkHG}
&h_{\beta,\sigma}(t_n)= \beta\frac{t_n}{\sigma}e^{\frac{-\pi t^2}{\sigma^2}} \ , n = -N_h,...,N_h
\end{align}
where $\beta$ and $\sigma$ are amplitude and time parameters, respectively. $\sigma$ can be chosen so that effective time support of the $h_{\beta,\sigma}(t_n)$ is set to half of the minimum chip duration. $\beta$ should be chosen as $\beta=2/\sum_{-N_h}^{N_h}|h_{1,\sigma}(t_n)|$. On the result of convolution, discontinuities in phase can be detected robustly by using Recursive Least Squares (RLS) technique. 

%
%

%
%
%

Convolution of detected signal’s instantaneous phase with the function $h_{\beta,\sigma}(t_n)$ is equal to effectively smoothed derivation operation, and provides more apparent phase jumps, as illustrated in Fig. \ref{fig:pp2}. Outliers of the convolved phase are detected by RLS method \cite{chen2002paper} and thereby phase shift points are determined, as illustrated in Figure \ref{fig:pp2_2}. This procedure does not provide any output for phase changes in frequency modulated signals and does not contribute to distinguishing frequency and phase modulated signals. Also, phase jump levels are discretized and vectorized, therefore the second pre-processing input is obtained to take place in classification of phase-shifted signals.

\subsection{Convolutional Neural Network Model and Feature Fusion}
Convolutional Neural Networks are widely used in image processing related problems for the automatic feature extraction and classification purposes. Input is convolved with a set of filters that of each is specialized for the detection of different local patterns. These convolution filter weights are updated during training phase so that they can detect local similarities in the image better. At the last layer of the CNN, class probabilities are given.

The proposed CNN model, as can be seen in Fig. \ref{fig:model}, has two inputs of reassigned TFI of the signal, which is obtained by preprocessing of the signal, and discretized phase difference vector, which is determined by and RLS adaptive filter, and gives the modulation types as output.   
Frequency modulated signals in time-frequency image form enables recognition through convolutional neural networks as they are in the image form. For the first pre-processed input, feature extraction process is performed in deep neural network of 3 convolutional layers, as illustrated in Fig. \ref{fig:model}. In these layers, 8, 4 and 2 filters are used with the size of 5x5, 4x4 and 4x4, respectively. The filter sizes are selected so that the lowest local similarity of the TFIs can be learned by the CNN. The unusual pattern with decreasing filter numbers is explained in the last paragraph of this chapter. Max pooling of size 2x2 is performed with stride of 2x2 after each layer to reduce computation, thereby decreasing size. 

1-dimensional 3 layered convolutional neural network is implemented as second feature extraction step using vectors obtained by second pre-processing. In these layers, 8, 4 and 2 filters are used with size of 5x1, 4x1 and 4x1, respectively. Max pooling of size 2x1 with stride of 2x1 is performed after each layer to reduce computation and decrease size. Lastly, feature fusion is applied to the output neurons of both CNNs, by combining the both networks last layers of 5 neurons and passing them through 2 dense layers where classification is performed. When the feature fusion layer is applied to the last layers of the CNNs, and training is performed as a single network instead of two separate classifiers, the resultant network model learns to tolerate errors and weak points of the individual pre-processing methods by adjusting the weights of the extracted features and manages to obtain highly accurate results.

Lastly, in CNNs it is a common approach to increase the number of channels while decreasing layer sizes progressively with the purpose of preventing information loss \cite{simonyan2014very}. However, increasing number of channels also increases the required computation as well as the number of parameters needs to be learned. In the proposed technique, similar to sparsifying autoencoder structures \cite{goodfellow2016deep}, both the size of layers and the number of channels are decreased to prevent excessive growth in the number of parameters and to ensure reduction in the size of layers progressively. As a result, a CNN structure that can successfully generalize over limited set of training data is obtained. 

\section{Simulation Results}
To compare the performance of the proposed method with the existing alternatives and to analyze its generalization capability of it at different SNR levels, two different sets are investigated. Types of modulations used in these scenarios, which are generated based on \cite{pace2009detecting}, are given in Table \ref{senaryolar}. Proposed FF-CNN is implemented in Python using Tensorflow library. For each changing number of training samples, constant number of 100 validation and 500 test samples are used per class. In addition, for some of the classes, data gathered from field measurements are also included in the test set (\%10-20 per class). Training is performed in batches of 128. All of these training, validation and test sets are chosen as mutually exclusive, in other words, network is tested on a set that it has not seen during training phase. Training is performed 3 times per scenario, and the weights giving best validation performance are used for the test samples to calculate the classification accuracy. Categorical cross-entropy is used as the loss function, and ADAM solver \cite{kinga2015method} is preferred for optimization, which combines the benefits of RMSProp and AdaGrad techniques.

\begin{figure}[t]
	\centering
	{
		{\includegraphics[width=3.5in]{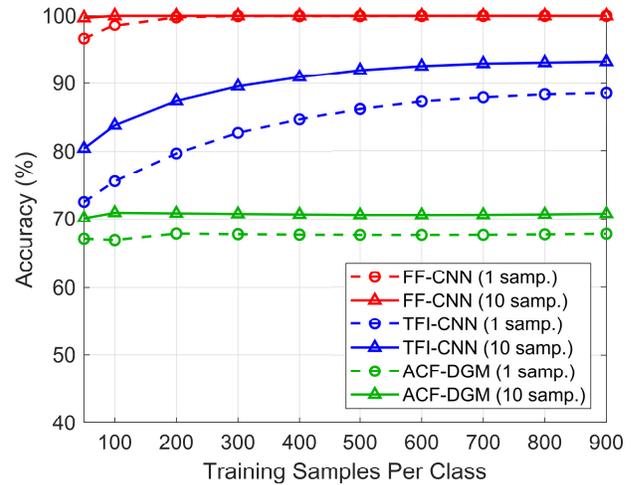}\label{fig:comp2}}
		
		\caption{Comparison of the proposed FF-CNN technique with two highest performing alternatives in 7 class case at 10 dB SNR} 
		\label{fig:comp}}
\end{figure}

In simulation scenarios, synthetic pulses with varying PW values from $(2-25)$ $\mu s$ are generated at 100 Mhz sampling rate at 5 and 10 dB SNR levels. Since the typical lowest SNR value for an EW system to detect a radar pulse in real-time is about 10 dB, the chosen values for SNR provide realistically challenging test cases. Pulses that has periodic frequency modulations (ramp, triangular and sinusoidal FM), are generated such that at least one period is present in $x(t)$. Stepped modulations are generated with at least 0.4 $\mu s$ chip duration. Number of frequency steps for Frank, P1 and P2 coded pulses are selected uniformly from $\{6,7,8\}$, and number of sub code in a code is selected uniformly from $\{36,49,67\}$ for P3 and P4 coded pulses. Number of segments are chosen uniformly from $\{4,5,6\}$ for T1 an T2 codes, and bandwidth of the intercepted signals are uniformly selected from $(5,10)$ Mhz for linear, ramp, triangular, sinusoidal FM, and T3-T4 coded pulses.
%
\begin{table}[t]
	\centering
	\caption{MODULATION TYPES USED AS CLASSES \\IN SIMULATION RESULT SETS}
	\renewcommand{\arraystretch}{1.2}
	\begin{tabular}[t]{C{4cm}C{1.7cm}C{1.7cm}}
		\toprule
		\multicolumn{1}{C{4cm}}{\textbf{7 Class Set}}  & \multicolumn{2}{c}{\textbf{23 Class Set}} \\
		\midrule
		Single Car. Mod. (SCM) & SCM & 8-PSK\\
		Linear FM & + Ramp FM & 16-PSK\\
		Costas-10 FM  & - Ramp FM & Frank Code\\
		Baker-13 PM & Sinusoidal FM & P1 Code\\
		QPSK & Triangular FM & P2 Code\\
		8-PSK & Costas-5 FM  & P3 Code\\
		16-PSK & Costas-7 FM  & P4 Code\\
		& Costas-10 FM  & T1 Code\\
		& Barker-3 PM & T2 Code\\
		& Barker-7 PM & T3 Code\\
		& Barker-13 PM & T4 Code\\
		& QPSK & \\
		\bottomrule
	\end{tabular}
	\label{senaryolar}
\end{table}
The first set is used to compare proposed FF-CNN technique with currently highest performing alternatives (\cite{wang2016radar} ACF-DGM, \cite{wang2017automatic} TFI-CNN) to the best of our knowledge. This set is the same set as used in \cite{wang2017automatic}, except our set also includes some additional phase modulations (QPSK, 8-PSK, 16-PSK). Convolution and averaging filter sizes used in TFI-CNN are optimized for the input size of $128 \times 256$ by cross-validation for a fair comparison. Results are obtained by calculating classification accuracies for individual predictions and combined-10-predictions (denoted as "10 samp." in figures), with changing training set sizes per class. For combined-10-predictions case, predictions of 10 test samples per class is combined to make the final decision. Fig. \ref{fig:comp} and Table \ref{perf_table} indicate that proposed FF-CNN technique outperforms the highest performing alternative technique by up to 10-15\%. The reason of TFI-CNN performing badly is, random QPSK, 8-PSK, 16-PSK sequences do not have very distinctive time-frequency images. Fusing the features extracted from second pre-processing input with the ones extracted from TFI input, enables improved classification.

\begin{figure}[t]
	\begin{center} 
		{\includegraphics[width=3.5in]{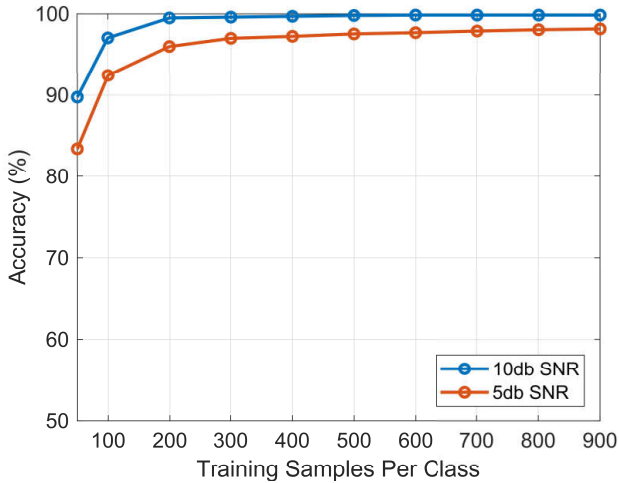}}
		
	\end{center} 
	
	\caption{Classification performance of the proposed FF-CNN technique for 23 class case at 5 and 10 dB SNR with different training set sizes} 
	\label{fig:14c}
\end{figure}

\begin{table}[t]
	\centering
	\caption{COMPARISON OF THE PROPOSED FF-CNN TECHNIQUE \\WITH TWO HIGHEST PERFORMING ALTERNATIVES \\IN 7 CLASS CASE AT 10 DB SNR}
	\label{perf_table}
	\renewcommand{\arraystretch}{1.2}
	\begin{tabular}[t]{C{2.6cm}C{2.4cm}C{2.4cm}}
		\toprule
		& \multicolumn{2}{c}{\textbf{Classification accuracies for}} \\
		& \multicolumn{1}{C{2.4cm}}{100 pulse per class} &\multicolumn{1}{C{2.4cm}}{ 900 pulse per class}\\
		\midrule
		\textbf{FF-CNN (1 pulse)} & \textbf{98.65\%} & \textbf{100.00\%} \\
		\textbf{FF-CNN (10 pulse)} & \textbf{100.00\%} & \textbf{100.00\%} \\
		TFI-CNN (1 pulse) & 75.57\% & 88.62\% \\
		TFI-CNN (10 pulse) & 83.83\% & 93.23\% \\
		ACF-DGM (1 pulse) & 67.10\% & 67.79\% \\
		ACF-DGM (10 pulse) & 70.10\% & 70.81\% \\
		\bottomrule
	\end{tabular}
\end{table}
The second simulation scenario set with 23 classes is used to test the scalability of the proposed FF-CNN technique over large number of classes at different SNR levels. As Fig. \ref{fig:14c} and Table \ref{süre} suggest, the proposed method is able to successfully classify 23 classes at 5 dB SNR without in a need of class specific classifier which makes this method feasible for any type of frequency/phase intra-pulse modulation including pseudo-random phase codes and radar-embedded communication \cite{blunt2010intrapulse} signal modulations.

Table \ref{süre} illustrates time analysis of the method for 24-class scenario. The method’s pre-processing is performed on CPU while network trainings and tests are performed on GPU. Models and specifications of the CPU and GPU are Intel i5 4460 with 4 cores of 3.4 GHz, and Nvidia GTX 970 with 1664 CUDA cores of 1050 MHz, respectively. After an off-line training, FF-CNN technique requires about 42 ms for classification of a pulse, which makes it feasible for on-line processing in EW systems.
\begin{table}[t]
	\centering
	\caption{AVERAGE TIME AND PERFORMANCE ANALYSIS \\FOR 23 CLASSES (900 TRAINING SAMPLES PER CLASS)}
	\label{süre}
	\renewcommand{\arraystretch}{1.2}
	\begin{tabular}[t]{L{3.9cm}C{1.7cm}C{1.7cm}}
		\toprule
		& \textbf{5dB SNR} & \textbf{10dB SNR} \\
		\midrule
		\textbf{Pre-processing Time (Training)} & 481 s & 472 s \\
		\textbf{Network Time (Training)} & 584 s & 541 s \\
		\textbf{Pre-processing (Testing)} & 39 ms & 39 ms  \\
		\textbf{Network Time (Testing)} & 3 ms & 3 ms \\
		\textbf{Performance (1 Sample)} & 98.10\% & 99.83\% \\
		\textbf{Performance (10 Sample)} & 99.85\% & 100.00\% \\
		\bottomrule
	\end{tabular}
\end{table}

\section{Conclusions}
In this work, a feature fusion based convolutional neural network structure is proposed for automatic classification of frequency and phase modulation types in radar pulses using TFI of pulses and detected anomaly part on the instantaneous phase of the received signal.
Simulation results show that the proposed FF-CNN technique outperforms the highest performing alternatives by a significant margin, and it is scalable over broad range of classes. The proposed FF-CNN structure can be trained by synthetic data, alleviating the difficulty of obtaining field data on rare modulation types. As a follow up of the encouraging results of this study, neural network structures that are capable of finding parameter values as well as class probabilities will be investigated in future works.


%
%



%
%
%

\bibliographystyle{IEEEtran}
\bibliography{refs}

\end{document}